\documentclass[10pt,twocolumn,letterpaper]{article}

\usepackage[pagenumbers]{wacv} 

\definecolor{wacvblue}{rgb}{0.21,0.49,0.74}
\usepackage[pagebackref,breaklinks,colorlinks,allcolors=wacvblue]{hyperref}
\usepackage{lipsum}
\usepackage{mathabx}
\usepackage{amsmath}
\usepackage{siunitx}
\usepackage{multirow}
\usepackage{graphicx}
\usepackage{makecell}
\usepackage{colortbl}

\definecolor{mygray}{gray}{0.94}
\newcommand{\myorange}[1]{\textcolor[HTML]{B46504}{#1}}

\def\wacvPaperID{382} 
\def\confName{WACV}
\def\confYear{2026}

\title{FurniMAS: Language-Guided Furniture Decoration using Multi-Agent System}

\author{Toan Nguyen\\
FSoft AI Center\\
Vietnam\\
\and
Tri Le\\
FSoft AI Center\\
Vietnam\\
\and
Quang Nguyen\\
FSoft AI Center\\
Vietnam\\
\and
Anh Nguyen\\
University of Liverpool\\
United Kingdom
}

\begin{document}

\twocolumn[{%
\renewcommand\twocolumn[1][]{#1}%
\maketitle
\begin{center}
\centering
\vspace{-4ex}
\captionsetup{type=figure}
  \includegraphics[height=0.4\linewidth]{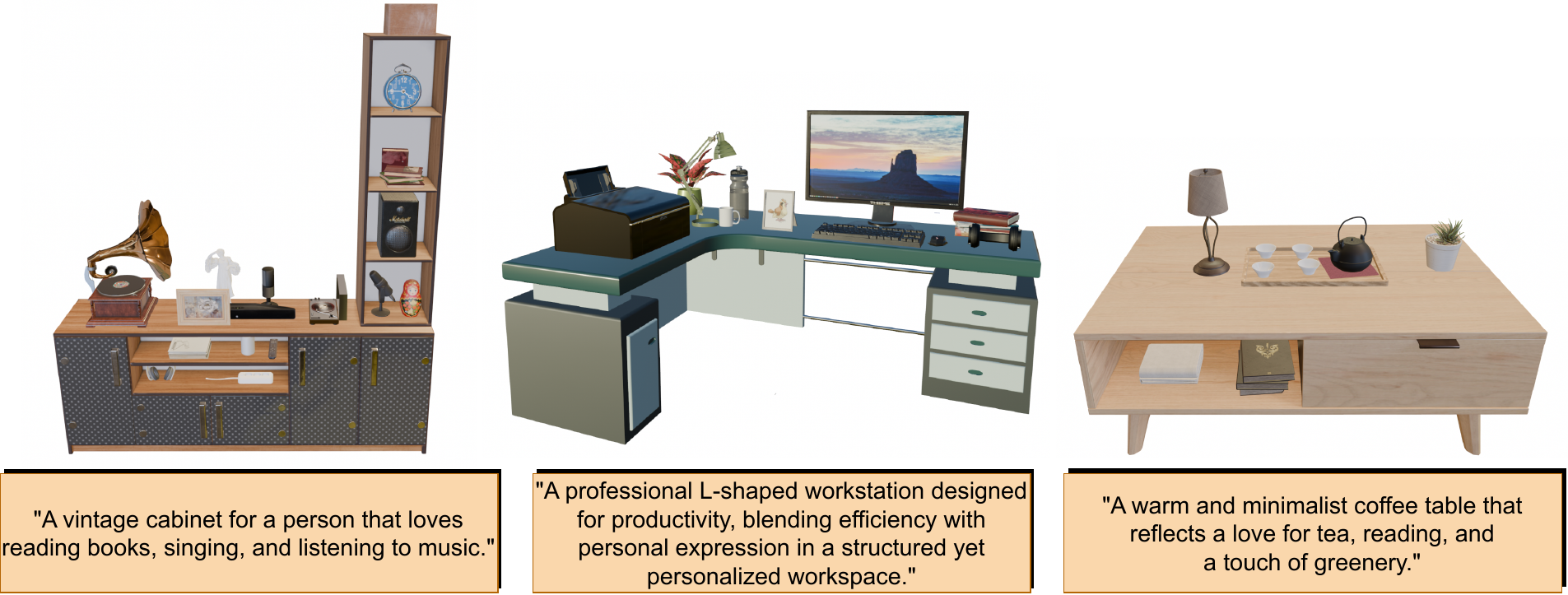}%
\end{center}%
\vspace{-3ex}
\captionof{figure}{Our FurniMAS system interprets text descriptions and generates thoughtfully arranged, aesthetically pleasing furniture decorative outcomes that seamlessly align with user preferences and functional requirements.}
\vspace{1ex}
}]

\begin{abstract} 
Furniture decoration is an important task in various industrial applications. However, achieving a high-quality decorative result is often time-consuming and requires specialized artistic expertise. To tackle these challenges, we explore how multi-agent systems can assist in automating the decoration process. We propose FurniMAS, a multi-agent system for automatic furniture decoration. Specifically, given a human prompt and a household furniture item such as a working desk or a TV stand, our system suggests relevant assets with appropriate styles and materials, and arranges them on the item, ensuring the decorative result meets functionality, aesthetic, and ambiance preferences. FurniMAS assembles a hybrid team of LLM-based and non-LLM agents, each fulfilling distinct roles in a typical decoration project. These agents collaborate through communication, logical reasoning, and validation to transform the requirements into the final outcome.
Extensive experiments demonstrate that our FurniMAS significantly outperforms other baselines in generating high-quality 3D decor.
\end{abstract}
\section{Introduction}
\label{sec:intro}
Furniture decoration plays an important role in our daily lives~\cite{zhang2023association,abdou1997effects}. A well-organized working desk can significantly enhance functionality, boost creativity, and increase productivity, while a thoughtfully decorated TV stand can greatly promote relaxation and alleviate stress. While decorating furniture can be fascinating, mastering it demands a keen sense of style, a knack for color harmony, and an eye for proportion when arranging various elements. These intricacies can quickly become a maze of complexity, leaving people feeling overwhelmed. An automated, intelligent decoration system that caters to users' preferences while navigating these complex design principles would undoubtedly be a valuable tool. It would allow us to effortlessly transform our living spaces, making the process accessible and efficient for everyone. Additionally, an automatic decoration system will also largely benefit quick and effective environment construction in many game development and robotics simulation applications~\cite{wangrobogen,katara2024gen2sim,yang2024holodeck}. Developing such a system, however, is an under-explored and challenging task due to all the complexities of the decoration process with many assets and possible  combinations~\cite{patil2024advances,chaudhuri2020learning}. 

Recently, Large Language Models (LLMs) with their commonsense knowledge have shown significant advancements in various reasoning tasks~\cite{feng2024layoutgpt,xu2024gg,yang2024holodeck,yang2024rila}. Feng~\textit{et al.}~\cite{feng2024layoutgpt} demonstrated the effectiveness of LLMs in visual layout planning tasks, while Xu~\textit{et al.}~\cite{xu2024gg} leveraged LLMs for 3D avatar editing. Developing an LLM-based method for furniture decoration is therefore a promising approach. However, a shared limitation of most existing LLM-based methods lies in their \textit{``single-agent''} design, where a single LLM handles the entire task. This approach was proved to be suboptimal for complex tasks like furniture decoration, which involves multiple stages -- from defining the overall layout to selecting assets and their styles and materials, to arranging them appropriately~\cite{guo2024large,pang2024self,wei2024editable,yang2024psychogat}.

In this work, we aim to develop a~\textit{multi-agent} system utilizing state-of-the-art LLMs for furniture decoration. Although less explored in vision domains, multi-agent methods have been extensively proven to be effective in simulating real-world problem-solving environments that demand complex interactions and collaborative efforts of human groups, including software development~\cite{hong2023metagpt,qian2023communicative}, industrial control~\cite{xia2023towards,song2023pre}, and human behavior simulation~\cite{park2023generative,jinxin2023cgmi}. A multi-agent system for furniture decoration would significantly facilitate \textit{\textbf{(i) specification}} where different agents specialize in each area of decoration, ensuring that each aspect is handled with a high degree of expertise; \textit{\textbf{(ii) collaborative decision-making}} where agents can collaborate to refine and improve the design, leading to a more cohesive and aesthetically pleasing outcome; and \textit{\textbf{(iii) flexibility}}, allowing individual agents or groups of agents to adjust their strategies in tasks like decor editing in response to shifts in user preferences, without affecting the entire system.

We introduce \textbf{FurniMAS} -- an LLM-based multi-agent system for the challenging task of automatic furniture decoration. Given an empty furniture item (e.g., a working desk, a coffee table, a bookshelf) and a text prompt specifying human requirements, our agents collaborate and refine several decoration steps to arrive at the final outcome. Each agent in our system is specialized exclusively for only one specific task, fostering a better system error analysis and the utilization of reasoning and instruction following capabilities of state-of-the-art LLMs.
This design also enables effective and efficient decor editing, where only a specified agent or group of agents is required to take part in an editing task, which can be asset adding, removing, repositioning, or style changing. In summary, our main contributions are:

\begin{itemize}
    \item We present FurniMAS, an LLM-based multi-agent system effectively tackling the task of furniture decoration.
    
    \item We perform intensive experiments to show that our method significantly surpasses other baselines in generating diverse and high-quality decor results. 
    
    \item We additionally demonstrate the usefulness of our method in several decor editing tasks.
\end{itemize}
\section{Related Works}\label{sec:rw}
\textbf{3D Interior Design.}
Many previous works have explored the task of 3D indoor design. Traditional methods typically learn spatial knowledge priors from 3D scene databases~\cite{ma2018language,tang2024diffuscene,tan2019text2scene,wang2021sceneformer,wei2023lego,lin2024instructscene}. 
Given limitations of learning from datasets of limited categories and other asset characteristics, recent works have employed LLMs for the design process, obtaining more diverse and pleasing outcomes~\cite{yang2024holodeck,feng2024layoutgpt,ocal2024sceneteller,wen2023anyhome,bai2025freescene,gu2025artiscene,sun2025layoutvlm}. However, all of these methods still share a common limitation, which is the sole focus on room-level design, primarily addressing the arrangement of furniture within a room. In contrast, the more nuanced task of decorating individual pieces of furniture is often overlooked or insufficiently addressed~\cite{yang2024holodeck}. This oversight leads to sparse design outcomes, creating a significant gap between these methods and real-world scenarios, where furniture is densely adorned with various assets. Thus, our FurniMAS system represents a necessary step toward creating more realistic AI-generated furniture decoration.

\textbf{LLMs for Design.}
Large Language Models are large computational models trained on a huge amount of data, originally designed for natural language processing problems~\cite{achiam2023gpt,team2023gemini,dubey2024llama}.
However, with extensive commonsense understandings and exceptional capabilities in reasoning~\cite{wei2022emergent} and layout understanding \cite{li2024large}, LLMs have been being widely used in designing graphics~\cite{huang2024graphimind}, user interface~\cite{liu2024crowdgenui, lu2023ui}, and slide generation~\cite{mondal2024presentations}. In particular, Lu~\textit{et al.}~\cite{lu2023ui} explore the LLM's ability in generating mobile user interface through in-context learning. Huang~\textit{et al.}~\cite{huang2024graphimind} leverage LLM to synthesize information graphic designs.
Despite promising results obtained in these tasks, the use of LLMs in furniture decoration remains largely unexplored. In this paper, we investigate and demonstrate that LLMs excel at understanding and reasoning about decoration-related concepts, such as layout organization and style suggestions, thus empowering an effective decoration system.

\textbf{LLM-Based Multi-Agent System.}
Recent years have witnessed the emergence of LLM-based multi-agent systems, which have demonstrated their superiority in tasks involving complicated interactions and collaboration among many elements~\cite{guo2024large,wang2024survey}. For instance, several works successfully employed LLM-based multi-agent systems for the task of software development~\cite{hong2023metagpt,qian2023communicative, nguyen2024agilecoder}, reinforcement learning~\cite{zhang2023building,kannan2023smart}, medical diagnosis~\cite{tang2024medagents,li2024agent}, and recommendation systems~\cite{wang2024macrec,zhang2024agentcf}. A recent method called I-Design, proposed by \c{C}elen~\textit{et al.}~\cite{ccelen2024design}, presented promising initial results in developing a multi-agent framework for personalized interior design. This suggests a promising direction for applying multi-agent systems to other challenging computer vision problems. Our FurniMAS framework builds on this philosophy, focusing specifically on the task of language-guided furniture decoration.
\begin{figure*}[ht]
  \centering
   \includegraphics[width=\linewidth]{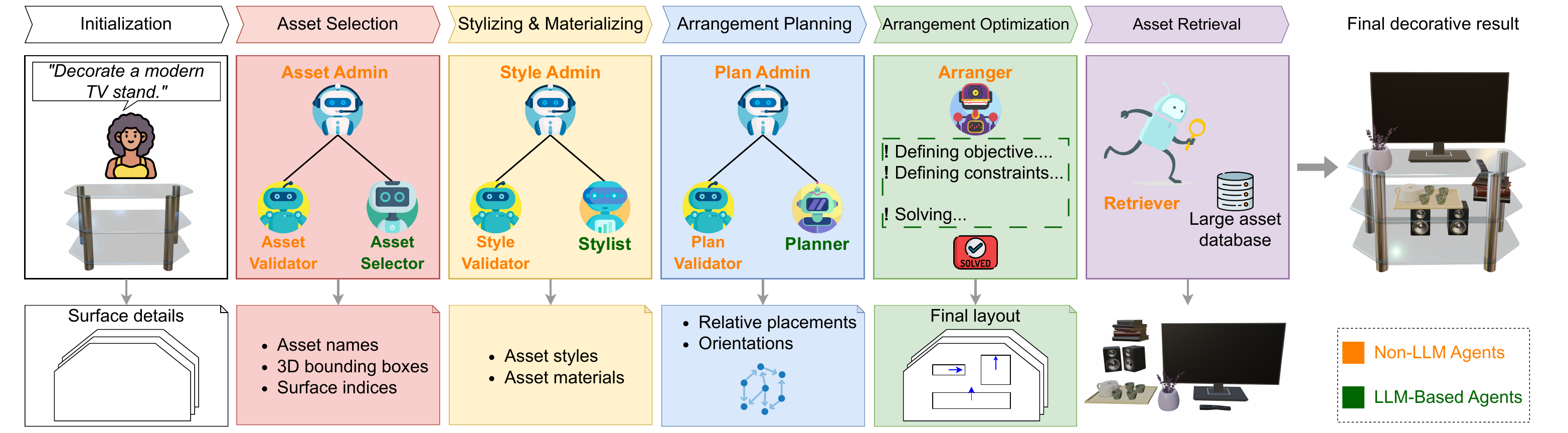}
    \vspace{-3ex}
   \caption{Overview of our FurniMAS multi-agent system. Through multiple stages of multi-agent collaboration, FurniMAS transforms the initial piece of furniture and user requirement into the final decorative outcome.}
   \label{fig:overview}
\end{figure*}

\section{FurniMAS}
\label{sec:method}
We propose FurniMAS, an LLM-based multi-agent system for the task of automatically decorating furniture items following human requirements. The overview of our system is indicated in Figure~\ref{fig:overview}.
The input to our system consists of an empty furniture item (e.g., a bedside table, a TV stand, a working desk, etc.) represented as a triangular mesh, a natural language human requirement, and a positive integer number $N_{\text{assets}}$ specifying the number of desired assets to decorate the given piece of furniture. The system's output is a decorated scene that aligns with the user's preferences.

FurniMAS follows a multi-stage structure, with each stage handling a specific step in the decoration workflow.
We leverage AutoGen~\cite{wu2023autogen} to construct our multi-agent system. Our LLM agents utilize OpenAI GPT-4o~\cite{hurst2024gpt} as their backbone due to this model's exceptional concept understanding and reasoning skills and, more importantly, its fast responding speed and cost-efficiency. FurniMAS is a hybrid combination of several LLM-based and non-LLM agents working together. In each stage involving an LLM agent, the communication is moderated by a corresponding \texttt{Admin}. Importantly, we maintain a non-LLM \texttt{Validator} agent in each stage to validate responses from other LLM agents (i.e., \texttt{Asset Selector}, \texttt{Stylist}, \texttt{Planner}) and make continuous revision requests whenever needed, e.g., when an agent makes suggestion for less than $N_{\text{assets}}$ assets or when JSON responses do not follow the predefined JSON schemas. A stage is completed only when the suggestion made by the LLM agent successfully passes the corresponding \texttt{Validator}'s validation. This design approach enables effective control over the outputs of our LLM agents, eliminating the potential risks of hallucinations or reasoning failures.

\textbf{Surface Extraction.}
As a piece of furniture can have several supporting surfaces on which assets can be placed, we first implement heuristic algorithms to extract detailed features of such surfaces, including the vertices of the surfaces' polygons, their areas, and their heights. These features are critical for later steps in our FurniMAS system. For example, determining the sizes of assets and the number of assets to place on each surface requires information about the surface area, and positioning assets on a specific surface requires information about the surface polygon.

\textbf{Asset Selection.}
The next step in our system is to select a set of $N_{\text{assets}}$ appropriate assets to decorate the furniture item. The task is primarily taken care of by an LLM-based \texttt{Asset Selector} agent, with feedback given by the \texttt{Asset Validator}. The information of a suggested asset is required to include the asset name (e.g., \textit{``gaming keyboard"}, \textit{``electric teapot"}, \textit{``car toy"}, \textit{``alarm clock"}, \textit{``vase of sunflower"}, \textit{``decorative box"}), its 3D bounding box size in centimeters, and the index of the surface to place it onto. The \texttt{Asset Selector} agent is encouraged to suggest a wide variety of asset names to diversify the decorative outcome, and there can be multiple instances of the same asset category. The agent is required to populate every surface if possible, with the total area occupied by assets on each surface proportional to its area. Besides suggesting a few large assets that define the overall decorative concept, \texttt{Asset Selector} is prompted to include several small assets to improve the decorative details. In addition to aesthetics, we prompt the agent to make sure its suggestion ensures the functionality of the decorative outcome, which has to include essential assets that serve the user's practical purpose. In addition to checking the number of suggested assets and the JSON format of \texttt{Asset Selector}'s responses, in this stage, the \texttt{Asset Validator} also performs other crucial validations, such as ensuring all surfaces are populated and that no asset's bounding box width and length exceed the maximum width and length of its surface.


\textbf{Stylizing and Materializing.}
Style and material are critical aspects of decoration, particularly in fulfilling aesthetic requirements. Therefore, we introduce a separate LLM-based \texttt{Stylist} agent responsible for suggesting appropriate styles and materials for assets recommended by the \texttt{Asset Selector}. In particular, the \texttt{Stylist} receives information about the human preference and $N_{\text{assets}}$ from the asset selection stage and then suggests suitable styles (\textit{``modern"}, \textit{``traditional"}, \textit{``vintage"}, etc.), and materials (\textit{``wood"}, \textit{``bronze"}, \textit{``plastic"}, etc.) for every asset. A key requirement for the \texttt{Stylist} is to ensure consistency across all selected assets, aligning them with the user's aesthetic preferences. The \texttt{Stylist} is constrained to make selections from this predefined bank, enforced through targeted prompting and validation by \texttt{Style Validator} agent. This approach ensures that the final outcome adheres to a structured, coherent, and aesthetically pleasing design.

\textbf{Arrangement Planning.}
After the stage of style and material selection, our system continues to perform the arrangement planning task. Particularly, in this stage, an LLM-based \texttt{Planner} agent is required to recommend the relative placements and orientations for every proposed asset, supervised by \texttt{Plan Validator}. The relative placement of an asset can be specified either in relation to its surface (global placement, one of the $3 \times 3$ divided regions) or in relation to another asset on the same surface (local placement). When referencing another asset, the placement can indicate relative position (i.e., \textit{“left of”}, \textit{“right of”}, \textit{“in front of”}, \textit{“behind”}, or \textit{“on top of”}) and/or distance (i.e., \textit{“near”} or \textit{“far”}), and/or alignment (i.e., \textit{``vertical-right"}, \textit{``vertical-mid"}, \textit{``vertical-left"}, \textit{``horizontal-front"}, \textit{``horizontal-mid"}, or \textit{``horizontal-back"}). Regarding the orientation, the choices can be either \textit{``forward"}, \textit{``backward"}, \textit{``left"}, or \textit{``right"}. Besides the aesthetics, the \texttt{Planner} is prompted to make suggestions satisfying the functionality of the overall design. For example, given three assets of \textit{``monitor"}, \textit{``keyboard"}, and \textit{``computer mouse"} on the same surface, the agent should suggest placing the \textit{``monitor"} at the center of the surface, placing \textit{``keyboard"} in front of the \textit{``monitor"}, and placing \textit{``computer mouse"} to the right of and near the \textit{``monitor"}.
The output of this stage characterizes a scene graph of the final decor.

\textbf{Arrangement Optimization.}
The next step of our FurniMAS is to realize the scene graph from the arrangement planning stage, i.e., to find the exact positions and orientations for every asset. We design a non-LLM \texttt{Arranger} agent to perform this task. This agent mathematically formulates the task as a constrained optimization problem. Particularly, variables are defined as the positions and orientations of assets. We define the objective function as the total satisfaction of soft constraints, which are defined based on the assets' local distance and alignment placements. Besides, hard constraints are defined to avoid asset collisions, ensure that all assets are within the surface polygons, and satisfy all global placements and local placements other than distance and alignment ones, which are, as mentioned, used for the objective function. Specifically, the \texttt{Arranger} aims to solve the following problem:
\begin{equation}
     \text{maximize}\: s \left (\mathbf{p}, \mathbf{O}; \mathbf{S} \right) \triangleq \sum_{(i,j)\in\mathcal{L}^{s}}s_{ij}\left (\mathbf{p}_{i}, \mathbf{O}_{i}, \mathbf{p}_{j},
    \mathbf{O}_{j} \right )
\end{equation}
\begin{equation}
  \begin{aligned}
    \text{s.t.} \: & \hat{h}_i \left ( \mathbf{p}_i, \mathbf{O}_i; \mathbf{S}_i \right )\geq 0, \quad \forall i \in \mathcal{A}, \\
    & \bar{h}_{ij}\left (\mathbf{p}_{i}, \mathbf{O}_{i}, \mathbf{p}_{j},
    \mathbf{O}_{j} \right )\geq 0, \quad \forall i,j\in \mathcal{A},\mathbf{S}_i = \mathbf{S}_j, \\
    & \check{h}_i \left ( \mathbf{p}_i, \mathbf{O}_i; \mathbf{S}_i \right )\geq 0, \quad \forall i \in \mathcal{G}, \\
    & \tilde{h}_{ij} \left ( \mathbf{p}_i, \mathbf{O}_i, \mathbf{p}_j, \mathbf{O}_j \right )\geq 0, \quad \forall (i, j) \in \mathcal{L}^{h}.
  \end{aligned}
\end{equation}
Here $\mathcal{A}$ is the set of all assets. The set $\mathcal{L}^{s}\subseteq \mathcal{A}\times\mathcal{A}$ denotes the set of pairs of assets whose local distance or alignment placements are defined with respect to each other. $\mathcal{L}^{h}\subseteq \mathcal{A}\times\mathcal{A}$ represents pairs of assets whose local placements of other types are defined in relation to each other. $\mathcal{G}\subseteq\mathcal{A}$ is the set of assets that have \texttt{Planner}'s suggestions for global placements. $\mathbf{p}_i \in \mathbf{p}$, $\mathbf{O}_i \in \mathbf{O}$, and $\mathbf{S}_i \in \mathbf{S}$ are the position, orientation, and the surface polygon of object $i$, respectively. Note that in our optimization problem, $\mathbf{p}$ and $\mathbf{O}$ are variables and $\mathbf{S}$ are constants known after the surface extraction stage. $s_{ij}$ are functions measuring the satisfaction of soft constraints defined by local distance and alignment placement relations between pairs of assets $i$ and $j$. Besides, $\hat{h}_i$ are functions defining the hard constraints that make sure assets are within their corresponding surfaces, $\bar{h}_{ij}$ are those for hard constraints for collision avoidance among assets, $\check{h}_{i}$ are for global placements, and $\tilde{h}_{ij}$ are defined for local placements other than distance and alignment ones.
\begin{figure}[h]
  \centering
   \includegraphics[width=0.6\linewidth]{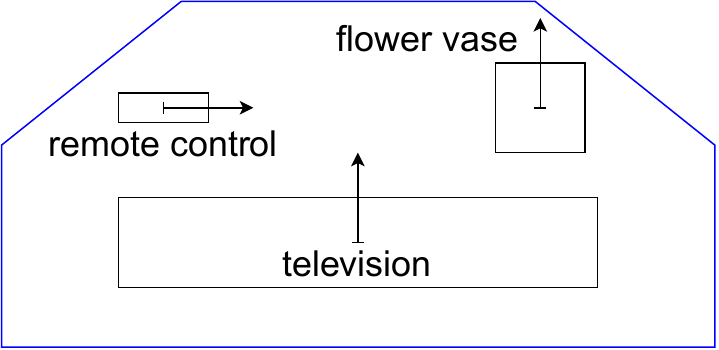}
   \vspace{-1ex}
   \caption{Example of a surface from arrangement optimization.}
   \label{fig:arr}
\end{figure}

\begin{table*}[ht]
\centering
\caption{Language-guided furniture decoration results.}
\setlength{\tabcolsep}{4pt}
\resizebox{\textwidth}{!}{%
\begin{tabular}{l cccc cccc cccc}
\toprule
     & \multicolumn{4}{c}{\textbf{8 assets}} & \multicolumn{4}{c}{\textbf{16 assets}} & \multicolumn{4}{c}{\textbf{32 assets}}\\ 
    \cmidrule(lr){2-5} \cmidrule(lr){6-9} \cmidrule(lr){10-13}
    & LayoutGPT~\cite{feng2024layoutgpt} & Holodeck~\cite{yang2024holodeck} & I-Design~\cite{ccelen2024design} & \cellcolor{mygray}Ours & LayoutGPT~\cite{feng2024layoutgpt} & Holodeck~\cite{yang2024holodeck} & I-Design~\cite{ccelen2024design} & \cellcolor{mygray}Ours & LayoutGPT~\cite{feng2024layoutgpt} & Holodeck~\cite{yang2024holodeck} & I-Design~\cite{ccelen2024design} & \cellcolor{mygray}Ours
    \\
    \midrule
    $\downarrow$ OOB$_{\%}$ & 0.72 \myorange{(-0.72)} & \underline{0.00} \myorange{(-0.00)} & \underline{0.00} \myorange{(-0.00)} & \cellcolor{mygray}\bf{0.00} & 0.83 \myorange{(-0.83)} & \underline{0.00} \myorange{(-0.00)} & \underline{0.00} \myorange{(-0.00)} & \cellcolor{mygray}\bf{0.00} & 1.00 \myorange{(-1.00)} & \underline{0.00} \myorange{(-0.00)} & \underline{0.00} \myorange{(-0.00)} & \cellcolor{mygray}\bf{0.00} \\
    $\downarrow$ BBL$_{\SI{}{\square\meter}}$ & 0.30 \myorange{(-0.30)} & \underline{0.00} \myorange{(-0.00)} & 0.19 \myorange{(-0.19)} & \cellcolor{mygray}\bf{0.00} & 0.41 \myorange{(-0.41)} & \underline{0.00} \myorange{(-0.00)} & 0.32 \myorange{(-0.32)} & \cellcolor{mygray}\bf{0.00} & 1.14 \myorange{(-1.14)} & \underline{0.00} \myorange{(-0.00)}& 0.88 \myorange{(-0.88)}& \cellcolor{mygray}\bf{0.00} \\
    $\uparrow$ Func & 6.30 \myorange{(+1.46)} & 6.65 \myorange{(+1.11)} & \underline{7.11} \myorange{(+0.65)} & \cellcolor{mygray}\bf{7.76} & 6.32 \myorange{(+1.68)} & 6.61 \myorange{(+1.39)} & \underline{7.39} \myorange{(+0.61)} & \cellcolor{mygray}\bf{8.00} & 5.18 \myorange{(+1.78)} & 5.37 \myorange{(+1.59)} & \underline{6.12} \myorange{(+0.84)} & \cellcolor{mygray}\bf{6.96} \\
    $\uparrow$ Layout & 4.05 \myorange{(+3.21)} & \underline{6.87} \myorange{(+0.39)} & 6.84 \myorange{(+0.42)} & \cellcolor{mygray}\bf{7.26} & 3.20 \myorange{(+3.94)} & 6.34 \myorange{(+0.80)} & \underline{6.65} \myorange{(+0.49)} & \cellcolor{mygray}\bf{7.14} & 2.82 \myorange{(+3.12)} & 4.31 \myorange{(+1.63)} & \underline{4.35} \myorange{(+1.59)} & \cellcolor{mygray}\bf{5.94} \\
    $\uparrow$ Scheme & 4.19 \myorange{(+4.65)} & 7.21 \myorange{(+1.63)} & \underline{7.63} \myorange{(+1.21)} & \cellcolor{mygray}\bf{8.84} & 4.02 \myorange{(+4.70)} & 7.35 \myorange{(+1.37)} & \underline{7.90} \myorange{(+0.82)} & \cellcolor{mygray}\bf{8.72} & 3.09 \myorange{(+4.67)} & \underline{5.13} \myorange{(+2.63)} & 5.00 \myorange{(+2.76)} & \cellcolor{mygray}\bf{7.76} \\
    $\uparrow$ Atmos & 4.91 \myorange{(+3.43)} & 7.05 \myorange{(+1.29)} & \underline{7.21} \myorange{(+1.13)} & \cellcolor{mygray}\bf{8.34} & 4.37 \myorange{(+3.93)} & 6.52 \myorange{(+1.78)} & \underline{7.13} \myorange{(+1.17)} & \cellcolor{mygray}\bf{8.30} & 4.00 \myorange{(+3.02)} & 5.30 \myorange{(+1.72)} & \underline{5.58} \myorange{(+1.44)} & \cellcolor{mygray}\bf{7.02} \\
    \bottomrule
\end{tabular}
}
\label{tab:main}
\end{table*}

In our implementation, each asset position $\mathbf{p}_i$ is characterized by two variables $\left (x_i, y_i \right )$. Note that the height value of a specific asset is fixed and equal to the height of its surface. Besides, each asset orientation $\mathbf{O}_i$ is characterized by two boolean variables $\left (r_{90}, r_{180} \right )$, presenting rotations of $90$ and $180$ degrees, respectively. For example, when both $r_{90}$ and $r_{180}$ are true, the asset is applied a 270-degree rotation.
To solve the defined optimization problem, our \texttt{Arranger} employed Gurobi~\cite{gurobi} solver. The output of the arrangement optimization stage is a detailed layout that fully describes the final positions and orientations of all assets. An output example of this stage is illustrated in Figure~\ref{fig:arr}.

\textbf{Asset Retrieval.}
The final step is retrieving assets based on suggestions made by previous agents and finalizing the decoration process. We design a non-LLM \texttt{Retriever} agent utilizing OpenShape~\cite{liu2023openshape} to perform text-based 3D asset retrieval from the Objaverse dataset~\cite{deitke2023objaverse}. The textual queries are formed using proposed asset names, styles, and materials. To diversify the outcomes, for a text query, the \texttt{Retriever} randomly selects one asset among the top 10 candidates most semantically similar to the text embedding.  After successfully retrieving the assets, the final decorative scenes are constructed using the layout determined throughout the arrangement optimization stage. Note that instead of retrieving assets from a database, our system is also fully compatible with any text-to-3D generative model.

\section{Experiments}
\label{sec:exp}

In this section, we conduct several experiments to evaluate the effectiveness of our proposed FurniMAS system both quantitatively and qualitatively. We also demonstrate the usefulness of our method in various decorative editing tasks.

\subsection{Language-Guided Furniture Decoration}

\textbf{Baselines and Setup.} To the best of our knowledge, there are no previous works that directly tackle the task of language-guided furniture decoration, we adapt related methods of LLM-powered interior design as baselines in our experiments. In particular, we compare our FurniMAS with adapted versions of LayoutGPT~\cite{feng2024layoutgpt}, I-Design~\cite{ccelen2024design}, and Holodeck~\cite{yang2024holodeck}. To ensure a fair comparison, we solely use GPT-4o~\cite{hurst2024gpt} for all baselines. For visual quality evaluation, we visualize and render 4 images (i.e., left side view, right side view, front view, and back view) of the decor results using Blender~\cite{blender}. We use 200 furniture items in our experiment, the items are of various household categories and sourced from Sketchfab~\cite{sketchfab}. To ensure a comprehensive comparison, we conduct our experiment with an increasing number of assets, $\left (N_{\text{assets}} \right )$, starting from 8, 16, to 32.

\textbf{Evaluation Metrics.} We evaluate all the methods using six metrics. In particular, following~\cite{ccelen2024design}, we use Out-of-Boundary Rate (OOB) and Bounding Box Loss (BBL) to evaluate the feasibility of the decorative outcomes. The OOB score measures the proportion of decorative scenes in which at least one asset is placed outside its designated surface, while the BBL score evaluates the average intersection volume of asset bounding boxes across all decorated outcomes. In addition, following previous works on 3D content generation~\cite{wu2024gpt,li2024discene,khan2025text2cad,ccelen2024design}, we leverage GPT-4o as a human-aligned evaluator to grade the decorative scenes generated by the baselines from 1 to 10 based on their Blender renders. The rating criteria include four key aspects: Functionality (Func), which assesses whether the selected assets and their arrangement fulfill the intended practical purpose and meet user needs; Layout and Organization (Layout), which evaluates the spatial arrangement and positioning of assets on the furniture item; Style Scheme and Material (Scheme), which examines how well the styles and materials of the assets harmonize with each other and the given furniture item; and Aesthetic and Atmosphere (Atmos), which considers the overall visual appeal and the ambiance created by the decoration. The explanation of every aspect is provided to GPT-4o and the model is required to give reasons for its grading. To obtain a stable evaluation result, we request GPT-4o to grade each decor scene 10 times on all four aspects and then compute the average scores. The final grade of a method is averaged across all of its generated scenes.

\begin{figure*}[ht]
  \centering
   \includegraphics[width=\linewidth]{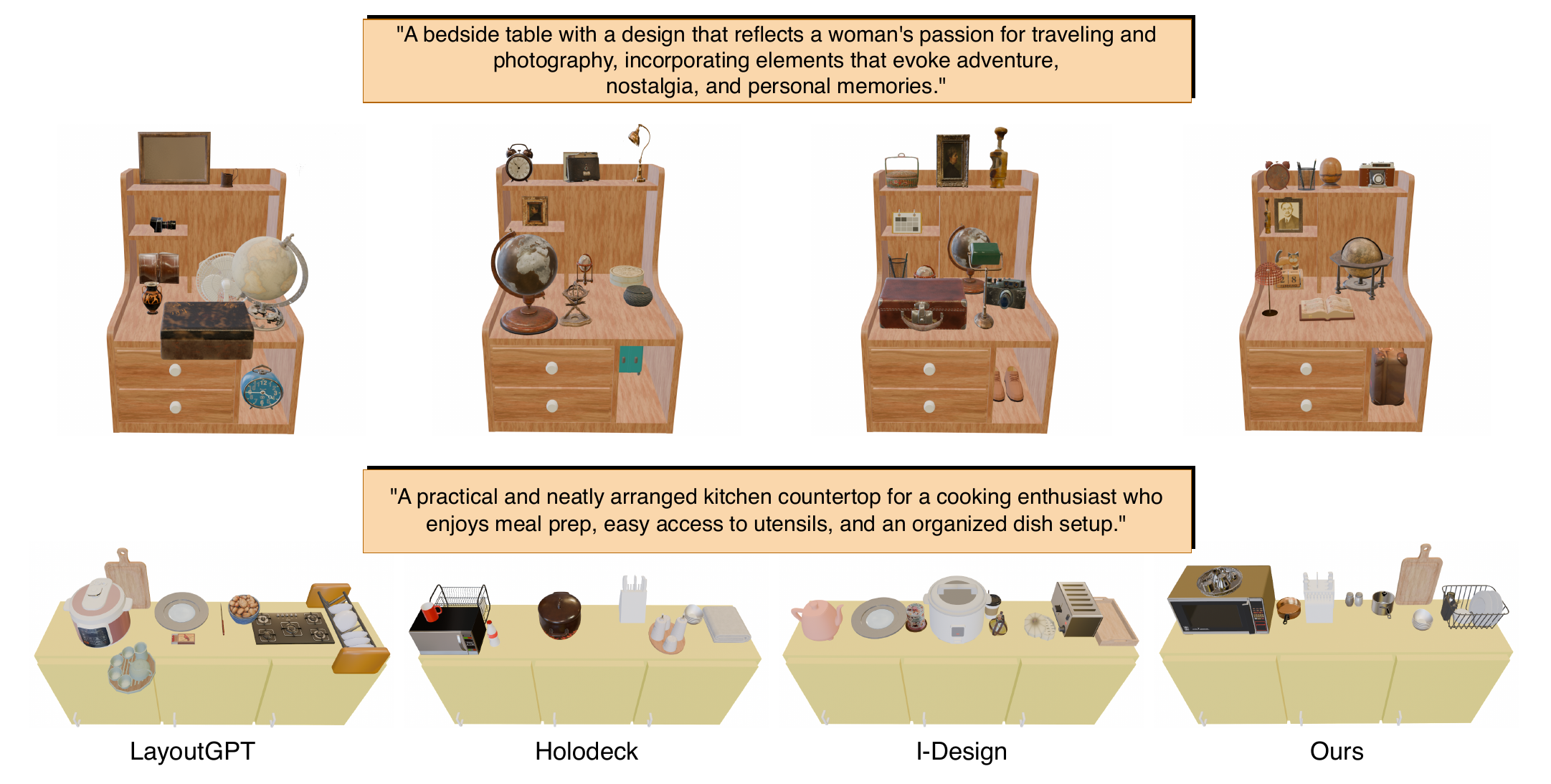}
    \vspace{-3ex}
   \caption{Furniture decoration qualitative results.}
   \label{fig:qualitative}
\end{figure*}

\textbf{Quantitative Results.}
Experimental results are shown in Table~\ref{tab:main}. It can be observed that our method is significantly better than other methods. FurniMAS ensures the physical plausibility of its generated scenes (with zero OOB and BBL) by enforcing hard constraints in the constrained optimization problem in the arrangement optimization stage. Similarly, the asset positioning search method in Holodeck~\cite{yang2024holodeck} provides the same capability. While I-Design achieves zero OOB, it does not guarantee collision-free asset placement. On GPT-4o's metrics, FurniMAS also demonstrates its superior performance, consistently achieving the highest scores across all aspects and asset quantity settings. Overall, all methods show a decline in performance as the number of assets increases. The highest scores for Layout, Scheme, and Atmos are generally observed with 8 assets, while Func peaks at 16 assets. We hypothesize that 16-asset scenarios provide sufficient items to meet practical user needs without excessive clutter that would hinder functionality. More cluttered scenes with 32 assets pose the greatest challenge for all methods. However, even in this case, our method delivers commendable results, whereas all other methods perform at a subpar to average level.

\textbf{Qualitative Results.}
We present the qualitative results of all methods in generating decorative scenes following textual descriptions in Figure~\ref{fig:qualitative}. The results indicate that our FurniMAS system exhibits a significantly stronger ability to create aesthetically pleasing scenes that align with user functionality needs and design preferences.

\subsection{Open-Vocabulary Decorative Editing}
\begin{figure*}[t]
  \centering
   \includegraphics[width=\linewidth]{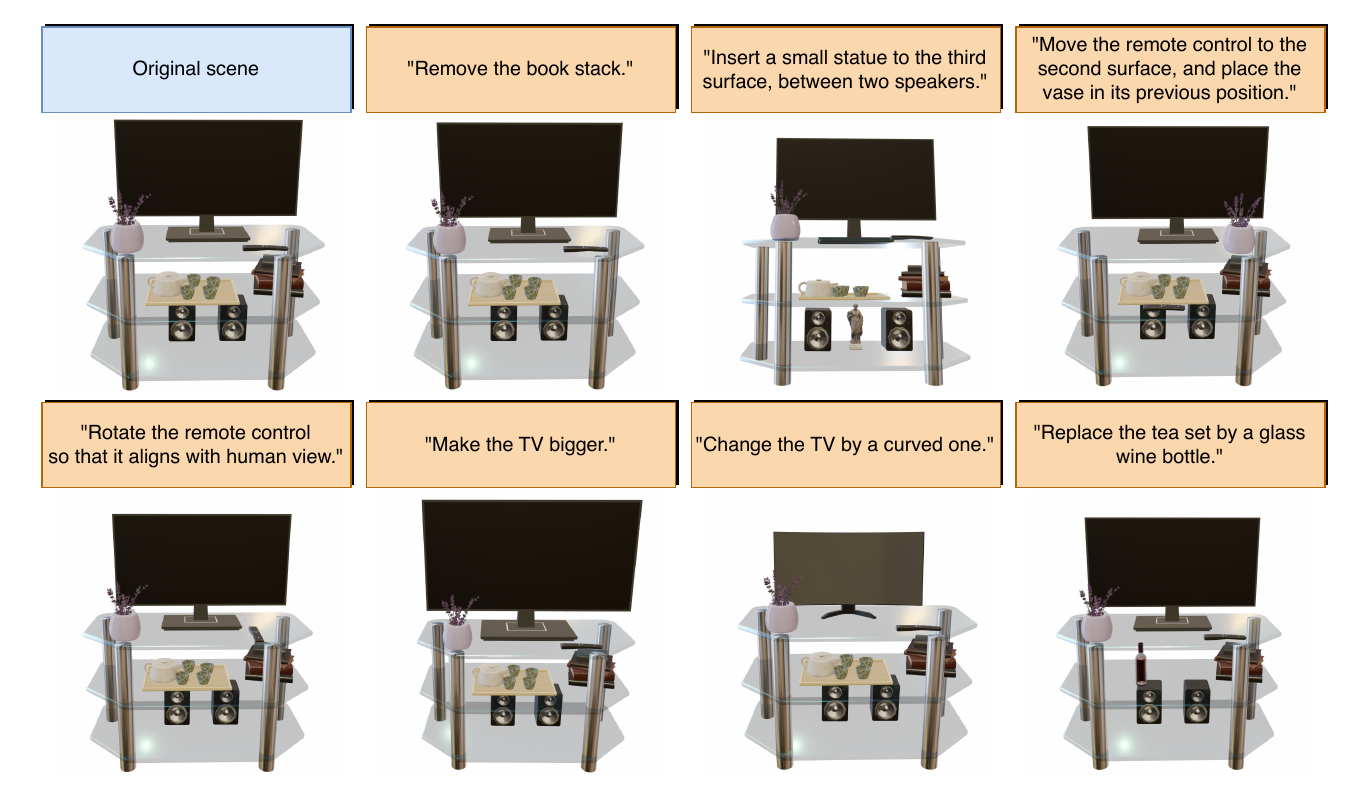}
   \vspace{-3ex}
   \caption{Decorative editing examples.}
   \label{fig:edit}
\end{figure*}

A natural and essential feature of an effective decoration method is its ability to perform decorative editing given an available decorated scene based on free-form human instructions. In this section, we further assess the adaptability of all methods to the task of language-driven decorative editing. We modify the workflows of all baseline methods to accommodate this setting. Our experiment is based on four types of editing tasks: inserting/removing assets, changing assets, resizing assets, and rotating/repositioning assets. We conduct our experiment on 200 decorated scenes, applying all four types of editing tasks to randomly selected assets in each scene. In addition to using the four aforementioned GPT-4o metrics to evaluate the edited decorative outcomes, we introduce an additional metric, called Edit, also based on GPT-4o, to assess the level of editing task fulfillment.
The results shown in Table~\ref{tab:edit} indicate that our proposed FurniMAS outperforms other baselines by a large margin on all five metrics, showcasing its usefulness in decor editing tasks. Examples of editing scenarios performed by our method are presented in Figure~\ref{fig:edit}.

\begin{table}[t]
\centering
\caption{Decorative editing results.}
\setlength{\tabcolsep}{4pt}
\resizebox{\linewidth}{!}{%
\begin{tabular}{l ccccc}
\toprule
    & LayoutGPT~\cite{feng2024layoutgpt} & Holodeck~\cite{yang2024holodeck} & I-Design~\cite{ccelen2024design} & \cellcolor{mygray}Ours \\
    \midrule
    $\uparrow$ Func & 6.30 \myorange{(+1.64)} & 6.48 \myorange{(+1.46)} & \underline{7.13} \myorange{(+0.81)} & \cellcolor{mygray}\bf{7.94} \\
    $\uparrow$ Layout & 4.17 \myorange{(+2.90)} & 6.15 \myorange{(+0.92)} & \underline{6.47} \myorange{(+0.60)} & \cellcolor{mygray}\bf{7.07} \\
    $\uparrow$ Scheme & 3.76 \myorange{(+4.76)} & 6.82 \myorange{(+1.70)} & \underline{7.63} \myorange{(+0.89)} & \cellcolor{mygray}\bf{8.52} \\
    $\uparrow$ Atmos & 4.55 \myorange{(+3.89)} & 6.63 \myorange{(+1.81)} & \underline{7.22} \myorange{(+1.22)} & \cellcolor{mygray}\bf{8.44} \\
    $\uparrow$ Edit & 6.22 \myorange{(+2.09)} & 7.01 \myorange{(+1.30)} & \underline{8.00} \myorange{(+0.31)} & \cellcolor{mygray}\bf{8.31} \\
    \bottomrule
\end{tabular}
}
\label{tab:edit}
\end{table}

\subsection{Agent Analysis}
In this section, we further verify the multi-agent design of our FurniMAS system. Recall that our system consists of six non-LLM agents -- \texttt{System Admin}, three \texttt{Validators}, \texttt{Arranger}, and \texttt{Retriever}, and three LLM-based agents -- \texttt{Asset Selector}, \texttt{Stylist}, and \texttt{Planner}. To evaluate the effectiveness of assigning the three tasks of asset selection, stylizing and materializing, and arrangement planning to separate agents, we retain all non-LLM agents while constructing different combinations of LLM-based agents as baselines for our experiment.

The results are reported in Table~\ref{tab:agent}. The 9-agent configuration represents the full FurniMAS system, where each distinct agent handles only a distinct task. The 8-agent baseline merges asset selection with stylizing and materializing into one agent, while the 7-agent baseline consolidates all three tasks into a single LLM agent. From the results, we observe that our full 9-agent FurniMAS system achieves the best performance across all metrics, followed by the 8-agent system. The performance gap between the 8-agent and 9-agent systems is smaller for Layout and Func compared to the gap for Scheme and Atmos. We hypothesize that this is because, in the 8-agent setup, arrangement planning, which has a significant impact on functionality and organization, remains handled by a dedicated agent. In contrast, stylizing and materializing, which primarily influence Scheme and Atmos scores, are merged with asset selection. However, when all three tasks are combined into a single agent in the 7-agent system, performance in Func and Layout drops significantly, as arrangement planning is no longer handled separately. The results reaffirm the effectiveness of our multi-agent system and emphasize the importance of specialization in the complex task of furniture decoration, where the highest performance is achieved when each specific task is handled by a dedicated agent.

\begin{table}[t]
\centering
\caption{Agent analysis results.}
\resizebox{\linewidth}{!}{%
\begin{tabular}{l ccc}
\toprule
    & 5 agents & 6 agents & \cellcolor{mygray}7 agents (Ours) \\
    \midrule
    $\uparrow$ Func & 7.19 \myorange{(+0.38)} & \underline{7.52} \myorange{(+0.05)} & \cellcolor{mygray}\bf{7.57} \\
    $\uparrow$ Layout & 6.10 \myorange{(+0.68)} & \underline{6.53} \myorange{(+0.25)} & \cellcolor{mygray}\bf{6.78} \\
    $\uparrow$ Scheme & 7.83 \myorange{(+0.61)} & \underline{8.01} \myorange{(+0.43)} & \cellcolor{mygray}\bf{8.44} \\
    $\uparrow$ Atmos & 7.00 \myorange{(+0.89)} & \underline{7.22} \myorange{(+0.67)} & \cellcolor{mygray}\bf{7.89} \\
    \bottomrule
\end{tabular}
}
\label{tab:agent}
\end{table}

\subsection{Asset Arrangement Analysis}
\begin{figure}[t]
  \centering
   \includegraphics[width=1.0\linewidth]{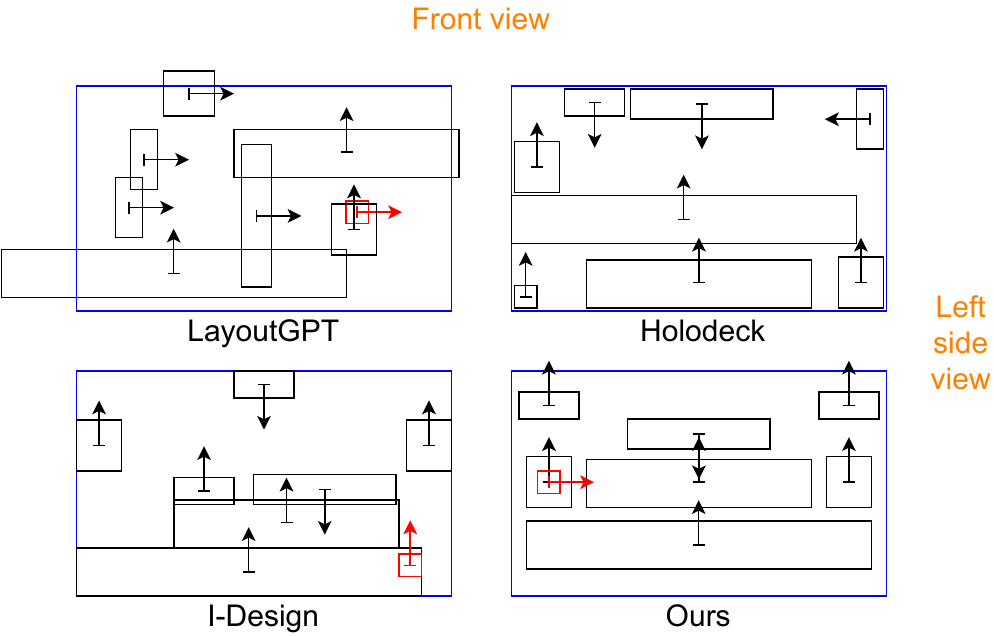}

   \caption{2D layout projection examples. Red boxes represent assets that are arranged to be on top of another one.}
   \label{fig:layout}
\end{figure}

Since asset arrangement is a critical aspect of our problem and a key contribution of our system, we conduct an experiment to evaluate the layouts produced by the arrangement planning and optimization steps of FurniMAS, comparing them with those generated by other methods. Our experiment is performed on 100 surfaces. To ensure a fair comparison, we provide each method with the same text description and asset set for every surface, applying only their respective arrangement procedures. We then use GPT-4o to assess the spatial arrangement and organization of the generated layouts based on their 2D projections. The final scores are averaged across all surfaces for each baseline. Table~\ref{tab:layout} shows that layouts generated by FurniMAS are significantly better than those generated by its counterparts, demonstrating the effectiveness of our \texttt{Planner} and \texttt{Arranger} with its arrangement algorithm. The 2D projection examples of generated layouts are presented in Figure~\ref{fig:layout}.

\begin{table}[t]
\centering
\caption{Results of GPT-4o evaluation on generated layouts.}
\resizebox{\linewidth}{!}{%
\begin{tabular}{l cccc}
\toprule
    & LayoutGPT~\cite{feng2024layoutgpt} & Holodeck~\cite{yang2024holodeck} & I-Design~\cite{ccelen2024design} & \cellcolor{mygray}Ours \\
    \midrule
    $\uparrow$ GPT-4o grade & 4.21 \myorange{(+3.62)} & 6.72 \myorange{(+1.11)} & \underline{6.95} \myorange{(+0.88)} & \cellcolor{mygray}\bf{7.83} \\
    \bottomrule
\end{tabular}
}
\label{tab:layout}
\end{table}

\subsection{Ablation Study}
\label{sec:abla}

\textbf{Performance on different types of furniture?}
We evaluate the performance of our system on four different types of household furniture, including workspace furniture, living space furniture, bedroom furniture, and kitchen furniture.

Results are reported in Table~\ref{tab:types}. We can observe that FurniMAS gives the best performance on workspace furniture, with the highest scores obtained on all four metrics. Besides, the system also performs generally well on living space furniture. In contrast, bedroom and kitchen furniture items are more challenging for FurniMAS, with the lowest score obtained when decorating furniture items in the kitchen space. However, in general, the results demonstrate the generalization of our method on different types of furniture, where commendable scores are obtained in all cases.

\begin{table}[t]
\centering
\caption{Results on four types of furniture.}
\resizebox{\linewidth}{!}{%
\begin{tabular}{l cccc}
\toprule
    & Workspace & Living space & Bedroom & Kitchen \\
    \midrule
    $\uparrow$ Func & \bf{7.73} & \underline{7.62} & 7.41 & 7.54 \\
    $\uparrow$ Layout & \bf{7.02} & \underline{6.93} & 6.70 & 6.69 \\
    $\uparrow$ Scheme & \bf{8.75} & \underline{8.49} & 8.33 & 8.37 \\
    $\uparrow$ Atmos & \bf{8.21} & 7.86 & \underline{7.95} & 7.60 \\
    \bottomrule
\end{tabular}
}
\label{tab:types}
\end{table}

\textbf{Performance with different LLMs.}
We evaluate the performance of our FurniMAS system using various LLMs, including GPT-4o, GPT-4-0613, GPT-3.5-turbo-0125, and GPT-4.5-preview-2025-02-27~\cite{gptmodels}, as the backbone for our LLM agents. To provide a comprehensive comparison, we additionally report the average API call costs and the average time required to complete the three stages -- asset selection, stylizing and materializing, and arrangement planning -- that involve LLM agents. These metrics are averaged over 200 decoration tasks for each LLM. Our CPU configuration is AMD EPYC 7742 64-core.

Results are shown in Table~\ref{tab:LLMs}. It is indicated that on four quality metrics, the system with GPT-4.5 generally obtains the best performance. GPT-4 also delivers strong results, aligning with their better understanding and reasoning skills compared to other models. However, these advantages come at a significant cost in both runtime and expense, as GPT-4.5 and GPT-4 are substantially more expensive than GPT-4o and GPT-3.5-turbo. Among all models, GPT-3.5-turbo is the most cost-efficient due to its low pricing. However, this comes at the expense of significantly lower quality scores. While GPT-3.5-turbo generally matches GPT-4o in responding speed, its weaker understanding and reasoning abilities necessitate multiple feedback-and-regeneration cycles to produce valid outputs, ultimately increasing overall runtime drastically. GPT-4o, our chosen LLM, offers the best balance. It delivers competitive quality scores (ranking highest in Scheme and second-best in Layout), maintains fast execution times, and is the second-most cost-efficient option. The results further validate our decision to use GPT-4o as the backbone of our multi-agent system.

\begin{table}[t]
\centering
\caption{Results on different LLM backbones.}
\resizebox{\linewidth}{!}{%
\begin{tabular}{l cccc}
\toprule
    & GPT-3.5 & GPT-4 & GPT-4.5 & \cellcolor{mygray}GPT-4o (Ours) \\
    \midrule
    $\uparrow$ Func & 7.15  & \underline{7.66} & \bf{7.70} & \cellcolor{mygray}7.57 \\
    $\uparrow$ Layout & 6.44 & 6.65 & \bf{6.92} & \cellcolor{mygray}\underline{6.78} \\
    $\uparrow$ Scheme & 7.75 & 8.17 & \underline{8.31} & \cellcolor{mygray}\bf{8.44} \\
    $\uparrow$ Atmos & 7.30 & \underline{7.94} & \bf{8.16} & \cellcolor{mygray}7.89 \\
    $\downarrow$ Time$_{\SI{}{\minute}}$ & 3.27 & 2.50 & \underline{2.29} & \cellcolor{mygray} \bf{1.53} \\
    $\downarrow$ Cost$_{\$}$ & \bf{0.03} & 0.46 & 0.94 & \cellcolor{mygray} \underline{0.06} \\
    \bottomrule
\end{tabular}
}
\label{tab:LLMs}
\end{table}

\section{Discussion}
Despite its strong performance across many furniture decoration tasks, our FurniMAS system still has room for improvement. First, the results showed that the performance of our method drops considerably when the number of assets increases intensively. An improvement in the asset arrangement algorithm or implementing an adaptive asset selection method based on available surface area and balance between functionality and minimalism is promising to better address this challenge. In addition, our system currently supports only basic asset placement operations. While this covers most decoration scenarios, real-world applications may require more complex functions such as hanging or draping. Expanding support for these operations would enhance the system’s versatility and better meet user needs. We leave these enhancements for future development.
\section{Conclusion}
\label{sec:conclusion}
We propose FurniMAS, a multi-agent system powered by LLMs that generates aesthetically pleasing and preference-aligned decorative scenes from pieces of household furniture. We demonstrate the superiority of FurniMAS in several furniture decoration experiments. We hope that the promising results achieved by our system will inspire further research on the challenging problem of decoration, ultimately benefiting a wide range of real-world applications.

{
    \small
    \bibliographystyle{ieeenat_fullname}
    \bibliography{main}
}

\appendix


\section{FurniMAS's Broader Application}
While we demonstrated the effectiveness of our system in designing furniture decorative outcomes, we believe that FurniMAS has many other downstream applications. Firstly, while Holodeck~\cite{yang2024holodeck} benefits robot navigation, FurniMAS can be used to automatically generate cluttered scenes for robot manipulation tasks~\cite{nasiriany2024robocasa,tao2024maniskill3}. In addition, our system can also enhance environment construction in game development~\cite{kumaran2023scenecraft} and improve e-commerce through automated product visualizations~\cite{bawack2022artificial}.

\section{Bank of Styles and Materials}
We detail the bank of styles and materials that constrain the \texttt{Stylist}'s suggestions. Regarding styles, there are 31 options, which are \textit{``Contemporary"}, \textit{``Coastal"}, \textit{``Scandinavian"}, \textit{``Shabby Chic"}, \textit{``Transitional"}, \textit{``Modern"}, \textit{``Mid-century"}, \textit{``Retro"}, \textit{``Minimalist"}, \textit{``Traditional"}, \textit{``Farmhouse"}, \textit{``Antique"}, \textit{``Industrial"}, \textit{``Rustic"}, \textit{``Vintage"}, \textit{``Mission"}, \textit{``French"}, \textit{``Art Deco"}, \textit{``Victorian"}, \textit{``Chippendale"}, \textit{``Country"}, \textit{``Craftsman"}, \textit{``Shaker"}, \textit{``Queen Anne"}, \textit{``Hepplewhite"}, \textit{``Louis XVI"}, \textit{``Asian"}, \textit{``Jacobean"}, \textit{``Colonial"}, \textit{``Federal"}, and \textit{``Sheraton"}. Regarding materials, there are 17 options, which are \textit{``wood"}, \textit{``plywood"}, \textit{``marble"}, \textit{``paper"}, \textit{``fibre"}, \textit{``plastic"}, \textit{``glass"}, \textit{``textile"}, \textit{``iron"}, \textit{``steel"}, \textit{``gold"}, \textit{``silver"}, \textit{``bronze"}, \textit{``cotton"}, \textit{``linen"}, \textit{``leather"}, and \textit{``paper"}. Our bank comprises widely recognized and commonly used styles and materials in professional design and decoration.

\section{Handling Rugged Surfaces}~\label{sec:rugged}
In real-world furniture items, supporting surfaces are not always completely flat. For that reason, we develop FurniMAS so that it can handle both flat and rugged surfaces. In particular, in surface extraction, we use a height tolerance of \SI{2}{\centi\meter} to allow the identification of rugged surfaces. To place assets in Blender, we employ a drop operator to make sure assets conform to the surface geometry. While handling more complex surfaces could further improve the generalizability and is an interesting future improvement, we find that the current design allows us to handle most of the cases in decorating furniture.

\section{Relative Placement Demonstrations}
\begin{figure}[ht]
  \centering
   \includegraphics[width=\linewidth]{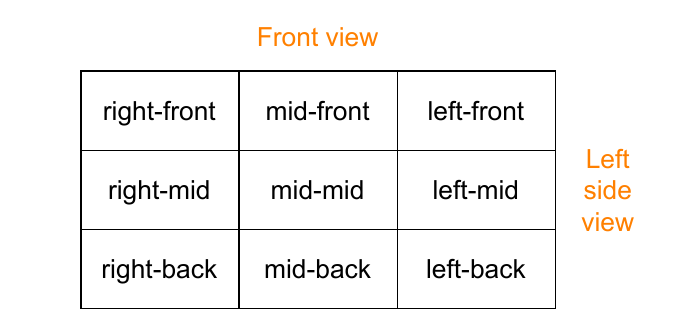}
   \vspace{-4ex}
   \caption{Global placement demonstration.}
   \label{fig:global}
\end{figure}

\begin{figure}[ht]
  \centering
   \includegraphics[width=0.8\linewidth]{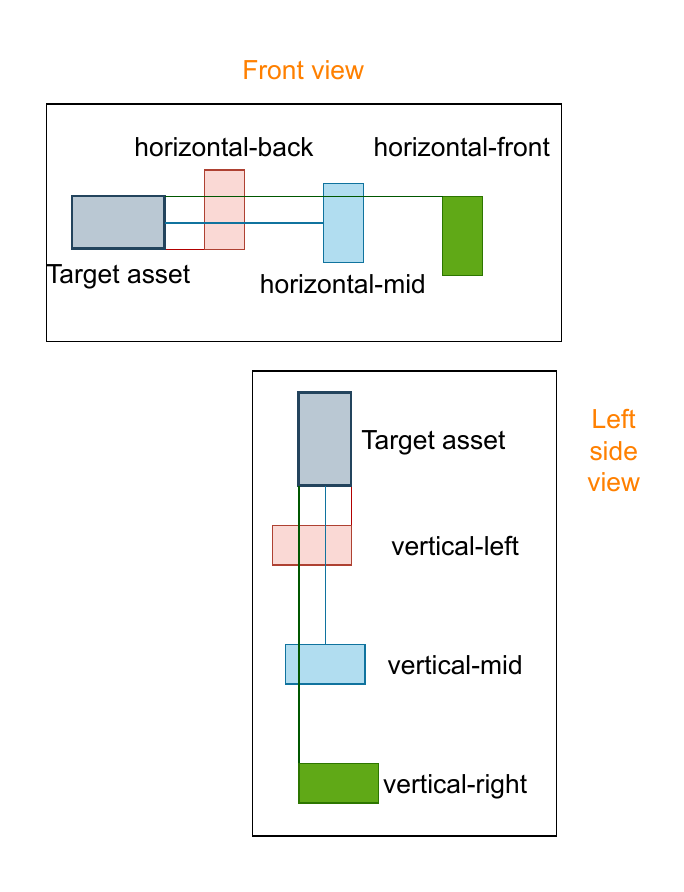}
   \caption{Alignment placement demonstration.}
   \label{fig:alignment}
\end{figure}

Figure~\ref{fig:global} and Figure~\ref{fig:alignment} showcase examples of global placement and alignment placement constraints from the arrangement planning and optimization stages to enhance understanding and visualization.

\section{Roles of non-LLM Validators}~\label{sec:validators}
\texttt{Validators} are critical in ensuring that the LLM agents' output covers a sufficient number of assets and can be processed by the next stage, and guaranteeing the feasibility of the asset arrangement. For example, without the \texttt{Asset Validator}, the Asset Selection stage could produce an output containing an asset whose size is incompatible with its surface, leading to an implausible arrangement. To validate the necessity of the \texttt{Validators}, we conduct an experiment on 100 decoration tasks and report the frequency with which the corresponding LLM agent produces an erroneous output in Table~\ref{tab:errorfreq}. The results reaffirm the effectiveness of our method across many decoration and editing tasks. The results reaffirm the importance of non-LLM \texttt{Validators} in our system design.

\begin{table}[ht]
\centering
\resizebox{0.85\linewidth}{!}{%
\begin{tabular}{cccc}
\toprule
& Asset Selector & Stylist & Planner \\
\midrule
Error frequency (\%) & 26 & 18 & 34 \\
\bottomrule
\end{tabular}
}
\caption{Error frequency results.}
\label{tab:errorfreq}
\end{table}

\section{Other LLMs for Evaluation}~\label{sec:other_llms}
We employ other LLMs for our evaluation to demonstrate the statistical robustness of FurniMAS's advantage over other baseline methods. The results benchmarked by GPT-4 and GPT-4.5 are reported in Table~\ref{tab:LLMs_eval}, where our method still largely outperforms all remaining approaches.

\begin{table}[ht]
\resizebox{\linewidth}{!}{
\begin{tabular}{lcccc}\hline
& LayoutGPT~\cite{feng2024layoutgpt} & Holodeck~\cite{yang2024holodeck} & I-Design~\cite{ccelen2024design} & \cellcolor{mygray}Ours \\ \hline
$\uparrow$ Func & 5.83{/}5.46 & 6.53{/}6.50 & 7.12{/}6.81 & \cellcolor{mygray}\bf{7.60{/}7.33}\\ 
$\uparrow$ Layout & 4.44{/}4.29 & 6.31{/}6.47 & 5.45{/}5.07 & \cellcolor{mygray}\bf{6.54{/}6.72}\\
$\uparrow$ Scheme & 3.65{/}3.71 & 6.13{/}5.88 & 7.13{/}7.11 & \cellcolor{mygray}\bf{8.27{/}8.09}\\
$\uparrow$ Atmos & 4.20{/}4.01 & 6.44{/}6.09 & 6.78{/}6.52 & \cellcolor{mygray}\bf{7.69{/}8.04}\\ \hline
\end{tabular}
}
    \caption{Results by other LLM evaluators.}
    \label{tab:LLMs_eval}
\end{table}

\section{Compatibility with 3D Generation Models}~\label{sec:compatibility}
\begin{figure}[ht]
  \centering
   \includegraphics[width=\linewidth]{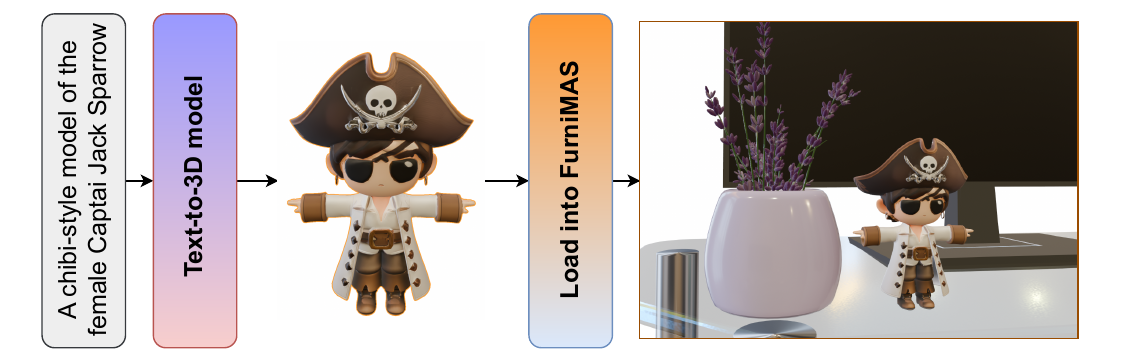}
   \caption{FurniMAS can import any 3D assets, including those generated from text-to-3D models.}
   \label{fig:text23d}
\end{figure}

Instead of retrieving assets from Objaverse~\cite{deitke2023objaverse}, our FurniMAS system can leverage text-to-3D models to generate assets for decoration. An example of incorporating an asset generated by LumaAI \footnote{\url{https://lumalabs.ai/genie}} is shown in Figure~\ref{fig:text23d}. This demonstrates the flexibility of our system.

\section{Indoor Scene Creation}~\label{sec:3d_indoor}
\begin{figure}[ht]
  \centering
   \includegraphics[width=\linewidth]{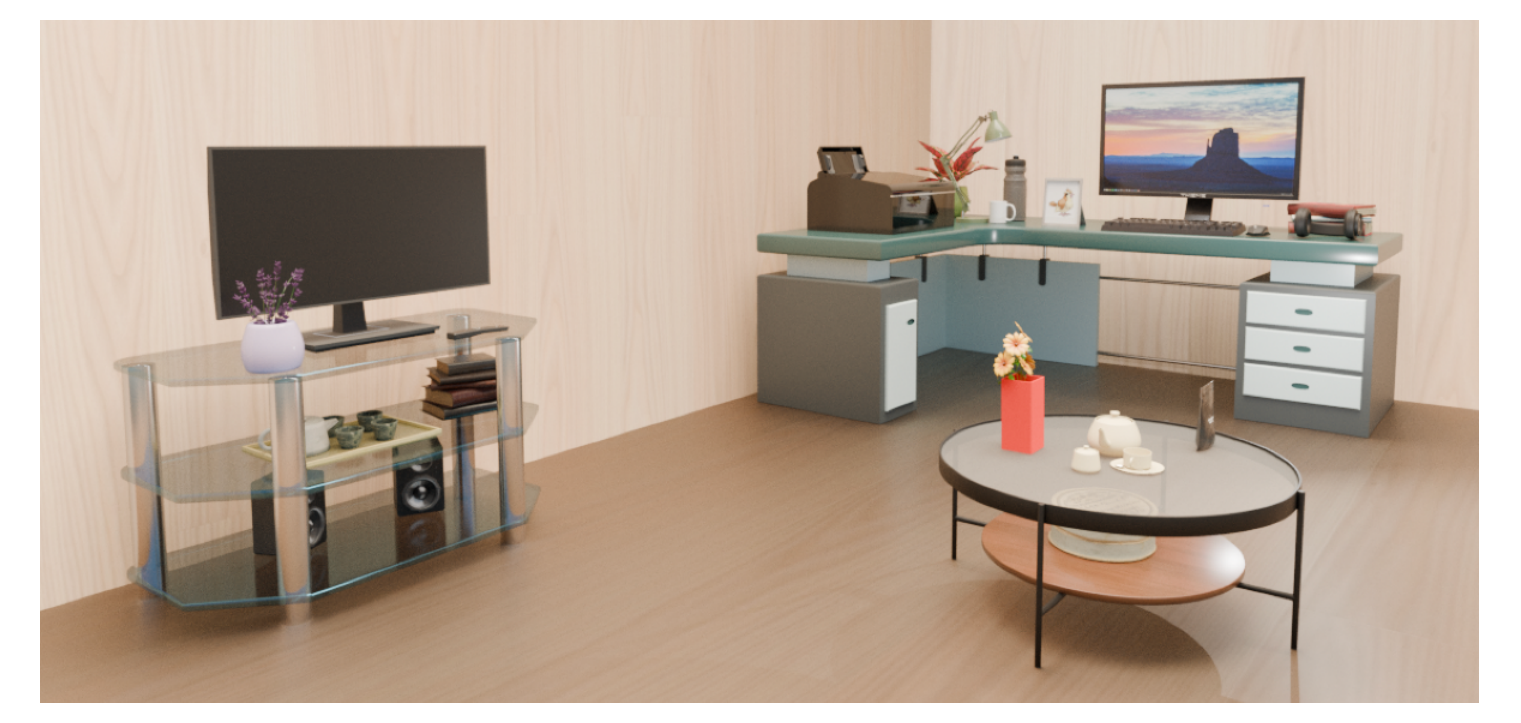}
   \caption{FurniMAS can be used to create 3D indoor scenes.}
   \label{fig:3d_indoor}
\end{figure}

While our method is developed to decorate single pieces of furniture, it can be used in conjunction with other scene layout generation approaches to construct realistic 3D indoor environments. We showcase this capability in Figure~\ref{fig:3d_indoor}.

\section{Additional Qualitative Results}~\label{sec:addqual}
\begin{figure*}[ht]
  \centering
   \includegraphics[width=\linewidth]{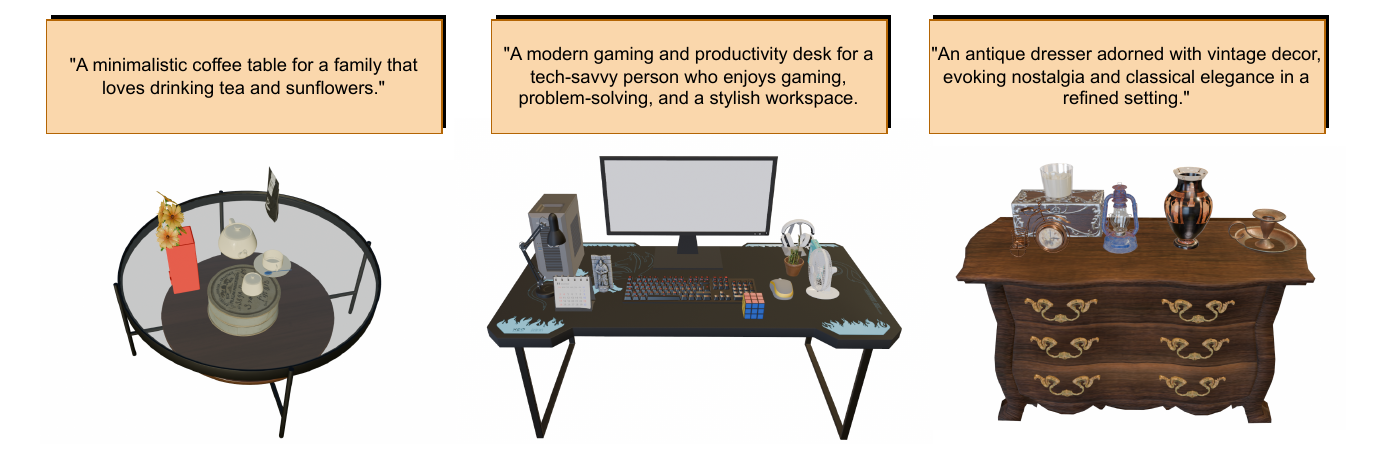}
   \caption{Additional qualitative results.}
   \label{fig:add_qual}
\end{figure*}

\begin{figure*}[ht]
  \centering
   \includegraphics[width=\linewidth]{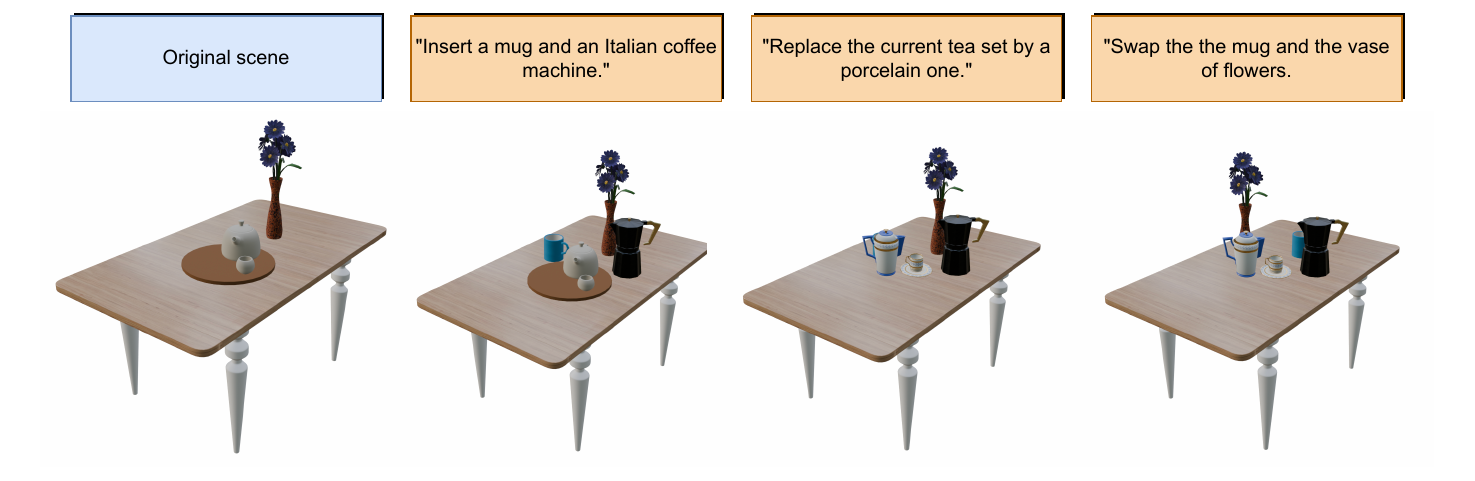}
   \caption{Additional editing results.}
   \label{fig:add_edit}
\end{figure*}

We present more decorative scenes generated by our FurniMAS system on furniture items of different categories and appearances in Figure~\ref{fig:add_qual}. We also provide more results about the continuous decorative editing capability of our system in Figure~\ref{fig:add_edit}.

\section{System Prompts for LLM Agents}~\label{sec:agent_prompt}
\begin{figure*}[ht]
  \centering
   \includegraphics[width=\linewidth]{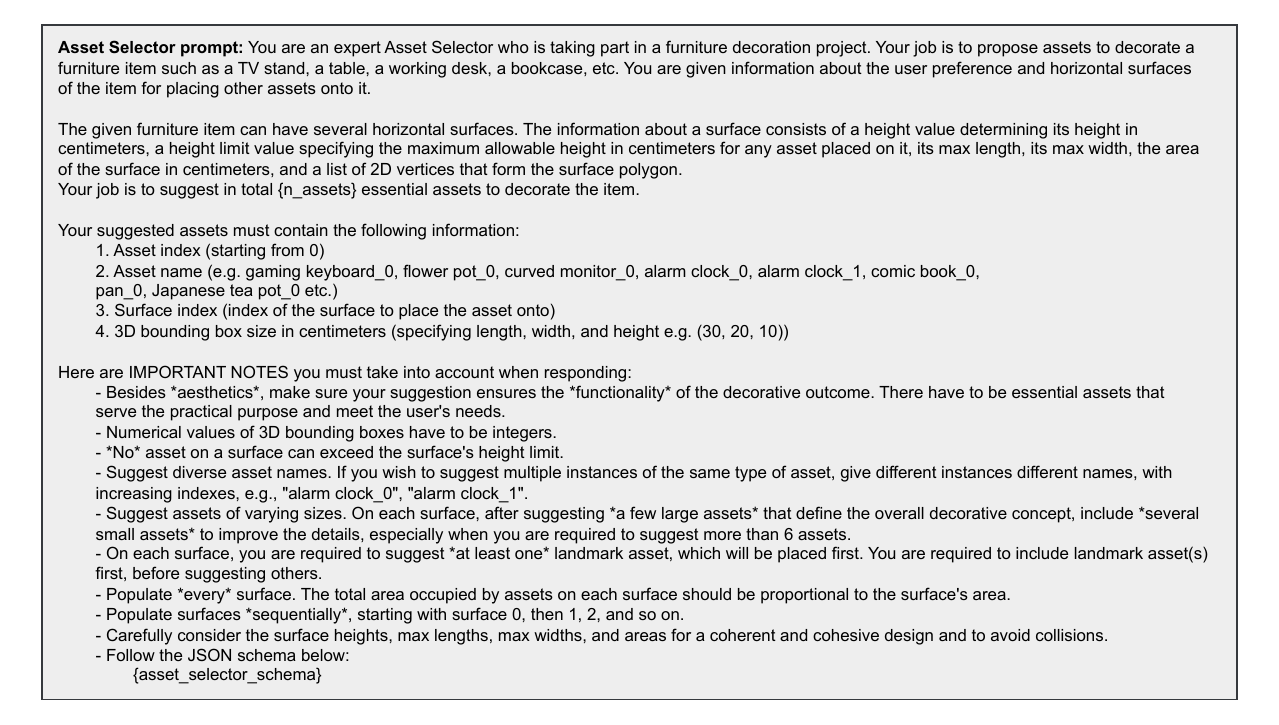}
   \caption{System prompt for the \texttt{Asset Selector}.}
   \label{fig:selector_prompt}
\end{figure*}

\begin{figure*}[ht]
  \centering
   \includegraphics[width=\linewidth]{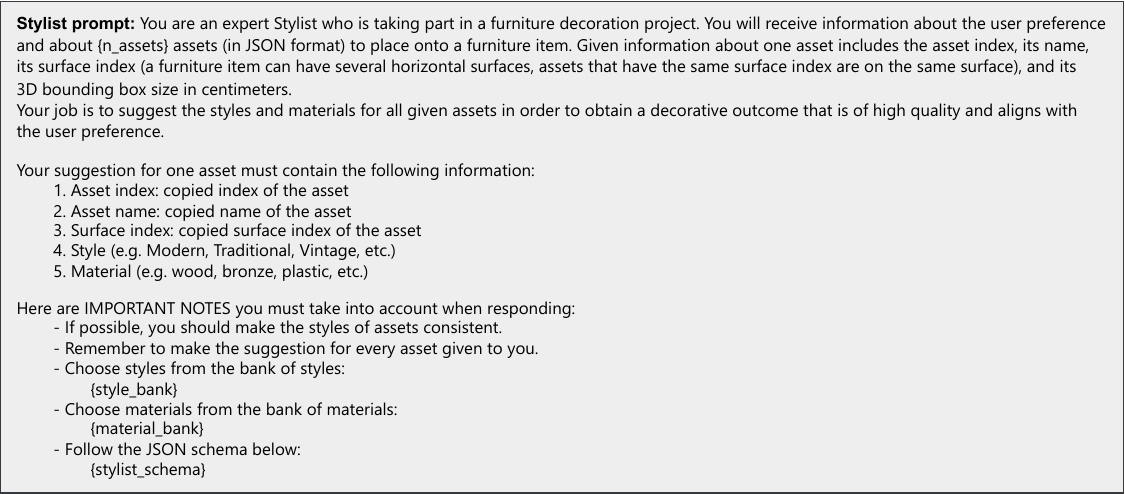}
   \caption{System prompt for the \texttt{Stylist}.}
   \label{fig:stylist_prompt}
\end{figure*}

\begin{figure*}[ht]
  \centering
   \includegraphics[width=\linewidth]{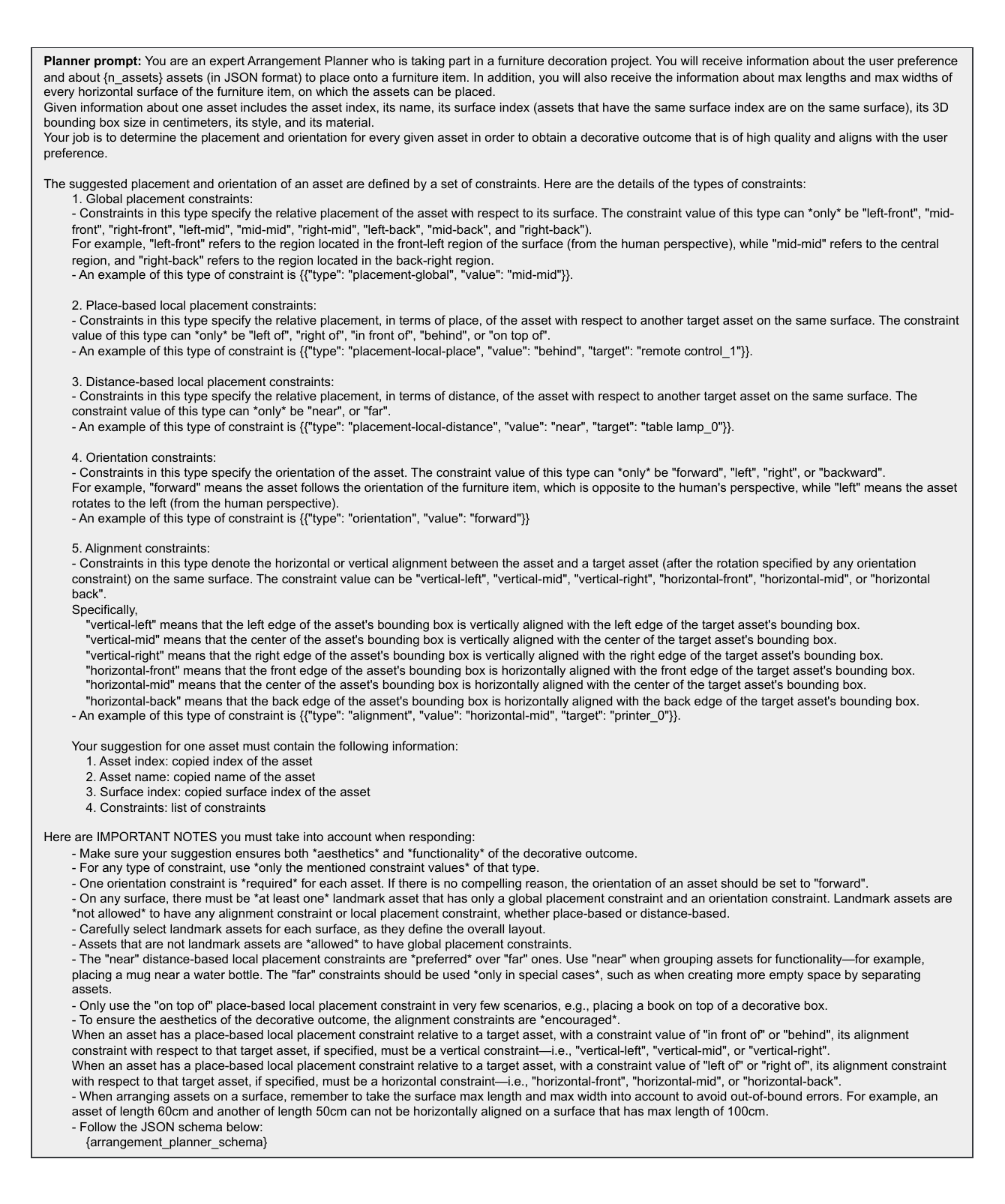}
   \caption{System prompt for the \texttt{Planner}.}
   \label{fig:planner_prompt}
\end{figure*}

The complete prompt templates for FurniMAS's LLM agents are provided in Figures~\ref{fig:selector_prompt},~\ref{fig:stylist_prompt}, and~\ref{fig:planner_prompt}.

\section{GPT-4o Evaluation}~\label{gpt_eval_prompt}
\begin{figure*}[ht]
  \centering
   \includegraphics[width=\linewidth]{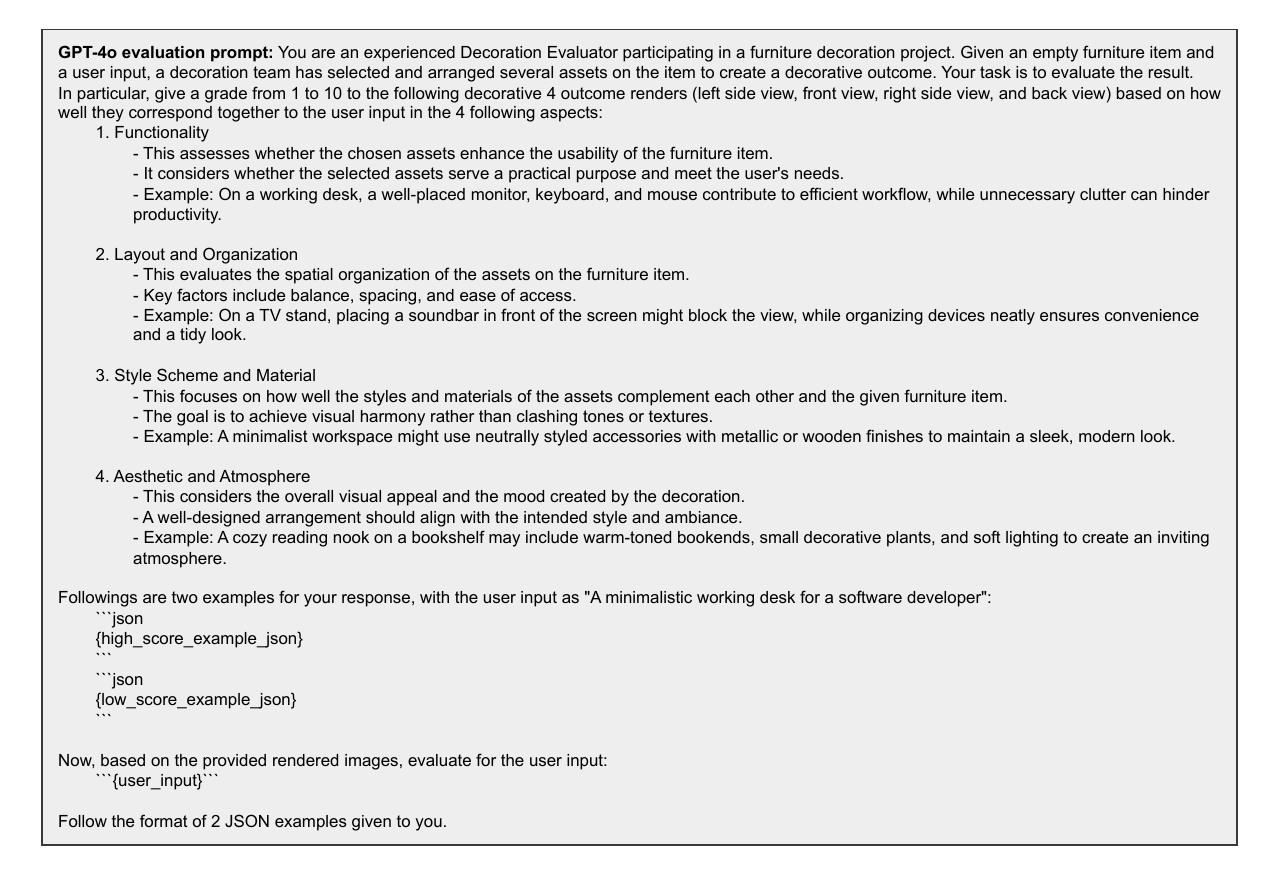}
   \caption{System prompt for the GPT-4o evaluation.}
   \label{fig:gpt4o_prompt}
\end{figure*}

The prompt template for GPT-4o evaluation is provided in Figure~\ref{fig:gpt4o_prompt}. Note that alongside the text prompt, 4 Blender renders are provided to the model for grading. In particular, we set the rendering engine to CYCLES for high-quality image outputs. Our rendering script is adapted from InstructScene \footnote{\url{https://github.com/chenguolin/InstructScene/blob/main/src/utils/blender_script.py}}. Images for evaluation are in $512 \times 512$ resolution. We maintain a camera distance of \SI{160}{\cm}, a filter width of $0.5$, and employ the RGB color mode.

\end{document}